\documentclass[conference]{IEEEtran}
\IEEEoverridecommandlockouts
\usepackage{cite}
\usepackage{amsmath,amssymb,amsfonts}
\usepackage{algorithmic}
\usepackage{graphicx}
\usepackage{textcomp}
\usepackage{xcolor}

\usepackage{acronym,url,makecell,placeins, amsmath,enumitem,graphicx,pdflscape,multirow}
\usepackage{algorithm2e}
\usepackage[numbers]{natbib}
\usepackage{booktabs}
\usepackage{xcolor}
\usepackage{pifont}
\newcommand{\xmark}{\ding{55}}%

\def\BibTeX{{\rm B\kern-.05em{\sc i\kern-.025em b}\kern-.08em
    T\kern-.1667em\lower.7ex\hbox{E}\kern-.125emX}}
\begin{document}

\title{Towards Improved Research Methodologies for Industrial AI: A case study of false call reduction}

\author{
\IEEEauthorblockN{Korbinian Pfab}
\IEEEauthorblockA{\textit{Department of Computer Science / Foundational Technology} \\
\textit{Helsinki University / Siemens AG}\\
Erlangen, Germany \\
0009-0009-8352-4216}
\and
\IEEEauthorblockN{Marcel Rothering}
\IEEEauthorblockA{\textit{Foundational Technology} \\
\textit{Siemens AG}\\
Erlangen, Germany \\
0000-0002-6028-9868}
}

\maketitle

\acrodef{AIabs}[AI]{artificial intelligence}
\acrodef{AI}[AI]{artificial intelligence}
\acrodef{PCBabs}[PCB]{printed circuit board}
\acrodef{SMTabs}[SMT]{surface-mounted technology}
\acrodef{AOIabs}[AOI]{automated optical inspection}
\acrodef{ML}[ML]{machine learning}
\acrodef{SMT}[SMT]{surface-mounted technology}
\acrodef{PCB}[PCB]{printed circuit board}
\acrodef{AOI}[AOI]{automated optical inspection}
\acrodef{MIS}[MIS]{manual inspection station}
\acrodef{V@S}[V@S]{volume reduction at target slip}
\acrodef{cV}[cV]{constrained volume reduction}
\acrodef{cAUC}[cAUC]{constrained area under curve}
\acrodef{AUC}[AUC]{area under the precision recall curve}
\acrodef{kNN}[kNN]{k-nearest-neighbor}
\acrodef{RFC}[RFC]{random forest classifier}
\acrodef{XGBoost}[XGBoost]{eXtreme Gradient Boosting}
\acrodef{BRFC}[BRFC]{balanced random forest classifier} 
\acrodef{DC}[DC]{dummy classifier}
\acrodef{AutoML}[AutoML]{automated machine learning}
\acrodef{ROC}[ROC]{receiver operating curve}
\acrodef{PRC}[PRC]{precision recall curve}

\begin{abstract}
Are current artificial intelligence (AI) research methodologies ready to create successful, productive, and profitable AI applications? This work presents a case study on an industrial AI use case called false call reduction for automated optical inspection to demonstrate the shortcomings of current best practices. We identify seven weaknesses prevalent in related peer-reviewed work and experimentally show their consequences. We show that the best-practice methodology would fail for this use case. We argue amongst others for the necessity of requirement-aware metrics to ensure achieving business objectives, clear definitions of success criteria, and a thorough analysis of temporal dynamics in experimental datasets. Our work encourages researchers to critically assess their methodologies for more successful applied AI research.
\end{abstract}


\section{Introduction}
\label{Introduction}
The rise of automation in manufacturing has brought significant advancements to production processes. However, are current \ac{AI} research methodologies ready to create successful, productive, and profitable \ac{AI} applications? Despite extensive research, the success of industrial \ac{AI} applications has not kept pace with other industrial automation technologies due to methodological weaknesses. 

In this work, we address these methodological flaws using a case study on false call reduction in \ac{AOI} of \acp{PCB}. \ac{AOI} systems, which use computer vision to inspect soldering quality, often produce a high number of false calls—incorrect classifications of non-defective \acp{PCB} as defective. These false calls consume valuable human resources in manual inspection stages.

Our study identifies seven prevalent weaknesses in related research on this topic and demonstrates their negative impacts experimentally. We highlight the necessity of using requirement-aware performance metrics over standard metrics, verifying assumptions about data distribution over time, and defining clear success criteria for experiments. By addressing these issues, we aim to challenge existing methodologies for evaluating and improving industrial \ac{AI} applications, ultimately enhancing their practical value and effectiveness. The key contributions of our paper are:
\begin{itemize}
    \item An analysis of common weaknesses in industrial \ac{AI} research based on related work on \ac{AI} applications to \ac{SMT} production
    \item A demonstration of measures to overcome the listed weaknesses such as the definition of requirement-aware metrics
    \item Delivering the first scientific results on the performance of \ac{ML} algorithms applied to false call reduction with published dataset and source code
\end{itemize}


\section{Background}
\label{Background}
In electronic production, there exist two main technologies for soldering \ac{PCB}s: through-hole technology and \ac{SMT}. For this work, we focus on \ac{SMT}. To ensure product quality and detect defects early and cost-efficiently, \ac{AOI} is commonly used directly after the \ac{SMT} soldering process. Images recorded by the \ac{AOI} are evaluated with computer vision algorithms to determine physical measurements such as displacement or rotation \citep{Michalska2020,Taha2014,Mar2011}. Inspection types, defined by the test engineer, specify the measurements and their acceptable values, which can change over time due to continuous improvement initiatives.

Based on the measurement defined by the inspection type, the \ac{AOI} classifies each soldering spot as defective or non-defective. Non-defective \ac{PCB}s move to the next process step, while defective ones go to a \ac{MIS} for manual inspection. However, many \ac{PCB}s classified as defective by the \ac{AOI} are later deemed non-defective (false calls) by the operator.

\begin{figure}[ht]
	\centering
		\includegraphics[width=0.48\textwidth]{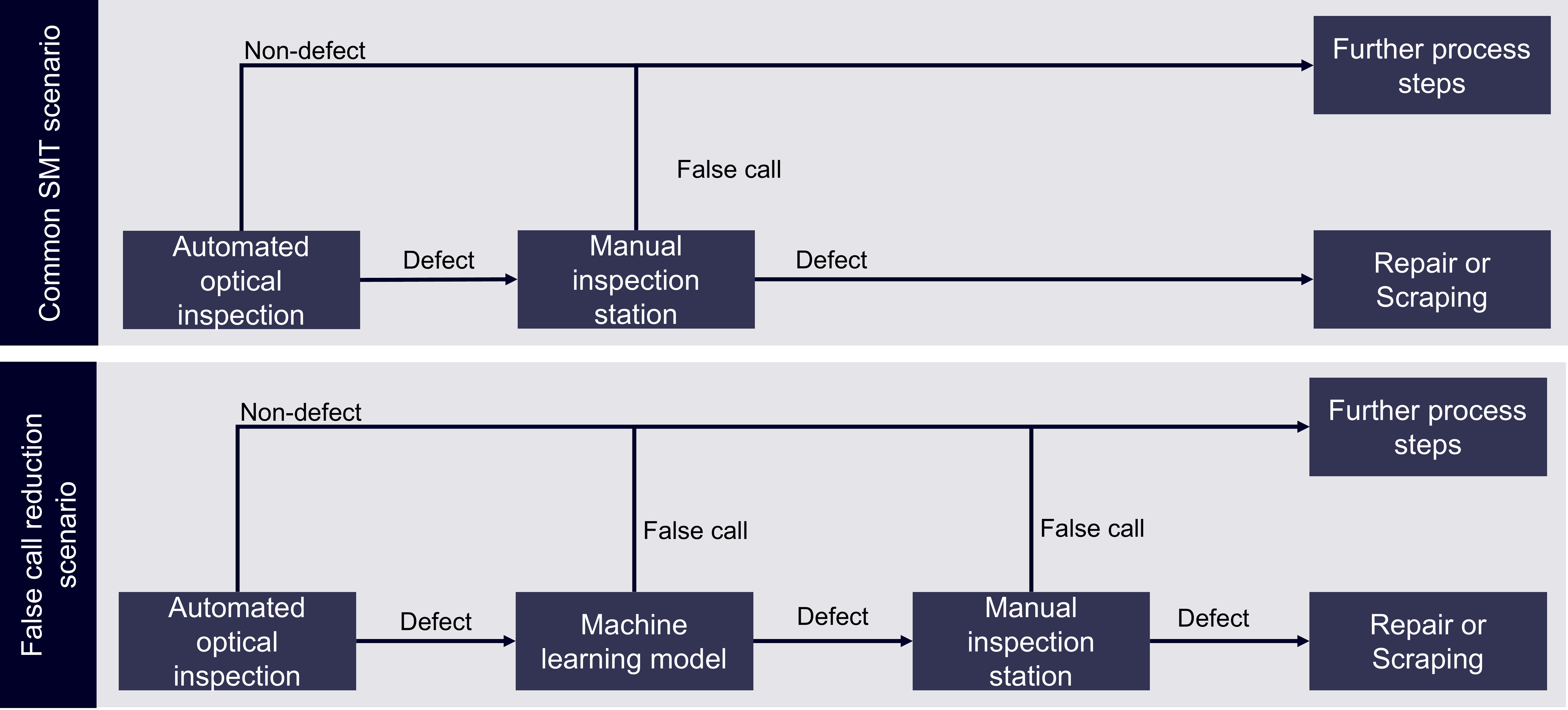}
	  \caption{Comparison of the common \ac{SMT} scenario and the enhanced false call reduction scenario}
   \label{workflow}
\end{figure}

Reducing false calls with \ac{ML} is a recent research topic. Different approaches introduce an \ac{ML}-based decision gate between the \ac{AOI} and \ac{MIS} to reduce the number of false calls to release operator capacity. Figure \ref{workflow} compares the common \ac{SMT} scenario with the enhanced false call reduction scenario.

If a truly defective board is wrongly classified as a false call by the \ac{ML} model and forwarded to further processes, it is considered a slip. Reducing the number of non-defective \ac{PCB}s at the \ac{MIS} through the \ac{ML} model is termed volume reduction. Incorrectly identifying a non-defective board as defective (false positives) decreases volume reduction. Volume reduction and slip rate are critical business metrics, with slips being significantly worse due to their negative impact on other production processes, while missed volume reduction reflects the current state without \ac{ML}.

\section{Related work}\label{related_work}

Two major approaches can be identified in related scientific work: using raw image data recorded by the \ac{AOI} or using measurement data extracted by the \ac{AOI}, soldering process, or soldering paste inspection. The classification object can vary from the quality of a whole \ac{PCB}, a single component, or a single soldering pin.

\citet{Lin2006} suggest a two-stage approach using image data of components. Features like the count of white pixels are calculated to classify a board as normal or one of three defective classes. They used 7768 non-defective and 90 defective samples, resulting in a highly imbalanced dataset. They used the false call rate and slip rate for evaluation but concluded that more efforts are needed. Their dataset is unpublished.

\citet{Jamal2022} use transfer learning with pre-trained convolutional neural networks, such as Xception \cite{XCeption}, on an image dataset of 4036 component images. They applied data augmentation methods and used accuracy as the evaluation metric. They achieved 91\% accuracy but admitted this might not be promising. Their dataset is unpublished.

\citet{Jabbar2019} used data from soldering paste inspection, labeling good, false call, and real defect. They downsampled the dataset to address extreme imbalance and tested tree-based models in a 10-fold cross-validation, achieving around 98\% accuracy and 97\% recall. An additional dataset of 2000 samples achieved 98.3\% accuracy, but classification level and class distribution details are unclear. Their dataset is unpublished.

\citet{Thielen2020} collected data on a component level from \ac{AOI} measurements, creating datasets of 1144 and 4264 samples. They evaluated neural networks, k-nearest neighbors, and random forest classifiers, with the latter performing best. They optimized thresholds to reduce slips to zero, but it remains uncertain if this would hold for future datasets. Their dataset is unpublished.

The discussed peer-reviewed related work contains seven common weaknesses:

\begin{enumerate}[label=\textbf{W\arabic*:},ref=\textbf{W\arabic*}]
    \item\label{W:A}\textbf{Lack of verification and reproducibility}
    
    The results cannot be verified or reproduced at all since neither the dataset nor the source code of the experiment is given. None of the discussed related work publishes its dataset or source code (cf. \cite{Jabbar2019, Jamal2022, Lin2006, Thielen2020}).
    \item\label{W:B}\textbf{Utilization of common models over advanced AutoML tools} 
   
    Instead of using state-of-the-art AutoML tools, a large topic are common models without naming certain requirements for that like explainability or inference time. None of the discussed related work considers AutoML approaches (cf. \cite{Jabbar2019, Jamal2022, Lin2006, Thielen2020}).
    
    \item \label{W:C}\textbf{Overemphasis on standard metrics and neglect of business impact}
    
    A strong focus is set on standard metrics, in specific accuracy, instead of metrics that consider the domain-specific use case requirements directly quantifying the business impact or standard metrics that incorporate potential trade-offs in a weighted manner. In \cite{Jabbar2019, Jamal2022} accuracy is used as main metric. \citeauthor{Jabbar2019} \cite{Jabbar2019} use accuracy, precision, F1-score, recall and Huber-loss. Only the recall is for this use case a truly meaningful metric. Nonetheless, for their final evaluation dataset only the accuracy value is given. In \cite{Jamal2022} only accuracy is considered.
    \item\label{W:D}\textbf{Lack of success criteria}
    
    The definition of requirements to classify the experiment as successful or not is neglected. Thus, the judgment of the results is often vague. 
    The related works \cite{Jabbar2019, Jamal2022, Lin2006} do not have a  success criteria. The authors of \cite{Thielen2020} define the goal of not introducing any slips, however, they do not define a goal for volume reduction.
    
    \item \label{W:E}\textbf{Inadequate handling of available information}
    
    The separation of information that is available at the time of the implementation and information that is not available is not strictly done. In  \cite{Thielen2020} the results for an adapted decision threshold are discussed. No methodology for determining such a threshold a priori is discussed and the work strongly implies that the decision threshold are determined a posteriori on the evaluated dataset which is not feasible in production. Also, performances based on decision threshold set on a specific dataset a posteriori cannot be seen as representative for additionally datasets.
    \item \label{W:F}\textbf{Neglecting temporal dynamics in the dataset}
    
    Temporal attributes of the dataset are not considered in the sense that there is no investigation done regarding distribution drifts in the dataset. While all of \cite{Jabbar2019, Jamal2022, Lin2006, Thielen2020} use different datasets for training and validation, it is not clear if the split was done randomly or sequentially. Only \citeauthor{Jabbar2019} \cite{Jabbar2019} mention for their final evaluation a new dataset, which implies a temporal split. However, none of the discussed related works gives a evaluation or analysis regarding similarity of their splits or temporal dynamics like data drifts in their dataset.
    
    \item \label{W:G}\textbf{Limited experiment variability due to single experiment runs}
    
    Experiments are just executed once for a certain random seed instead of evaluating multiple runs using different random seeds. None of the discussed related works shows the results of multiple runs (cf. \cite{Jabbar2019, Jamal2022, Lin2006, Thielen2020}).
\end{enumerate}

As the number of related work for this use case is limited, in Table \ref{tab:lit_comp} we present an analysis of extended related work to strengthen the point that the listed weaknesses are common in industrial \ac{AI} research. This analysis is based on the publications discussed in \cite{10168524}, which is a literature review of \ac{ML} application related to \ac{SMT}. Besides \textbf{W5}, all weaknesses are regularly present. However, to truly analyzing whether \textbf{W5} is present, one would require the source code used and the corresponding dataset, which is not the case due to \textbf{W1}.

\begin{table}[ht]
\centering
\renewcommand{\arraystretch}{1.5}
\caption{Analysis on what weaknesses are present in what related works on \ac{ML} applications to \ac{SMT}}
\begin{tabular}{cccccccccc}
    \hline
        \multirow{2}{*}{\makecell{Related\\work}} & \multicolumn{7}{c}{Weaknesses}\\
    \cline{2-8}
         & \textbf{W1} & \textbf{W2} & \textbf{W3} & \textbf{W4} & \textbf{W5} & \textbf{W6} & \textbf{W7}\\
    \hline
    \cite{Nian2018} & \checkmark & \checkmark & \checkmark & (\checkmark) & \xmark & \checkmark & \checkmark \\
    \cite{Chang2019} & \checkmark & \checkmark & \checkmark & \checkmark & \xmark & \checkmark & \checkmark \\
    \cite{Soto2019} & \checkmark & (\checkmark) & \checkmark & \checkmark & \xmark & \xmark & \checkmark \\
    \cite{SCHMITT2020} & \checkmark & \checkmark & (\xmark) & \checkmark & \xmark & \checkmark & (\checkmark) \\
    \cite{Khader2018} & \checkmark & - & \xmark & \xmark & \xmark & \checkmark & (\checkmark) \\
    \cite{KHADER2017} & \checkmark & - & \xmark & \xmark & \xmark & \checkmark & (\checkmark) \\
    \cite{PARVIZIOMRAN2019202} & \checkmark & \checkmark & \checkmark & \checkmark &\xmark & \checkmark & \checkmark\\
    \cite{PARVIZIOMRAN2019100} & \checkmark & \checkmark & \checkmark & \checkmark &\xmark & \checkmark & \checkmark\\
    
\end{tabular}

\label{tab:lit_comp}
\end{table}

\section{Methodology}
\label{methodology}

In our research, we give insights into the effects of common weaknesses in the research about industrial \ac{AI} applications on the example of false call reduction. Thereby, we utilize the dataset described and published in \citep{PfabData} and share our source code in \citep{PfabGit}. This dataset is chronologically ordered, tabular and consists of 77 columns including a timestamp and a label column featuring the labels \textit{false calls} and \textit{defect}. Furthermore, it partially consists of categorical features that we one-hot-encode and we drop the timestamp column for the modelling. The label classes have an extreme imbalance ratio of 99\%. 

\begin{figure}[ht]
	\centering
		\includegraphics[width=0.48\textwidth]{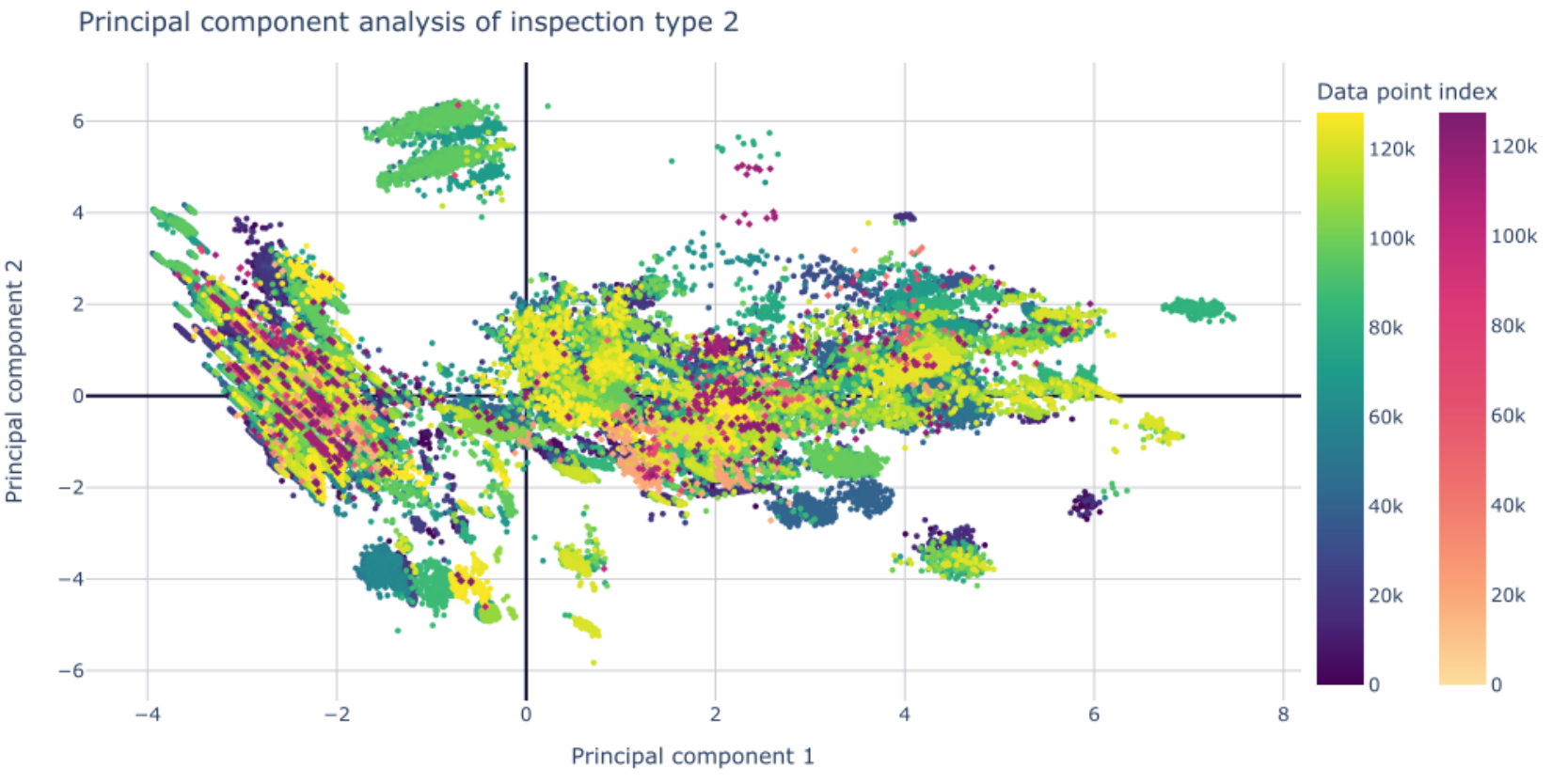}
	  \caption{Principal component analysis of the inspection type 2 of the dataset. The left color scale is for false calls and the right color scale is for true errors.}
   \label{PCA}
\end{figure}

Additionally, a time dependency of the dataset can be seen in Figure \ref{PCA}. It shows the output of a two-dimensional principal component analysis applied to the dataset and the color of the points are based on their classes and their indexes in the dataset. Multiple clusters can be identified which clearly have different colors indicating that a certain cluster did just appear in a certain period of time. More information about the dataset can be found in the data repository \citep{PfabData} and its corresponding publication.

This dataset follows the approach of using the extracted measurements of the \ac{AOI} machines on the soldering pin level.

\begin{figure}[ht]
	\centering
		\includegraphics[width=0.38\textwidth]{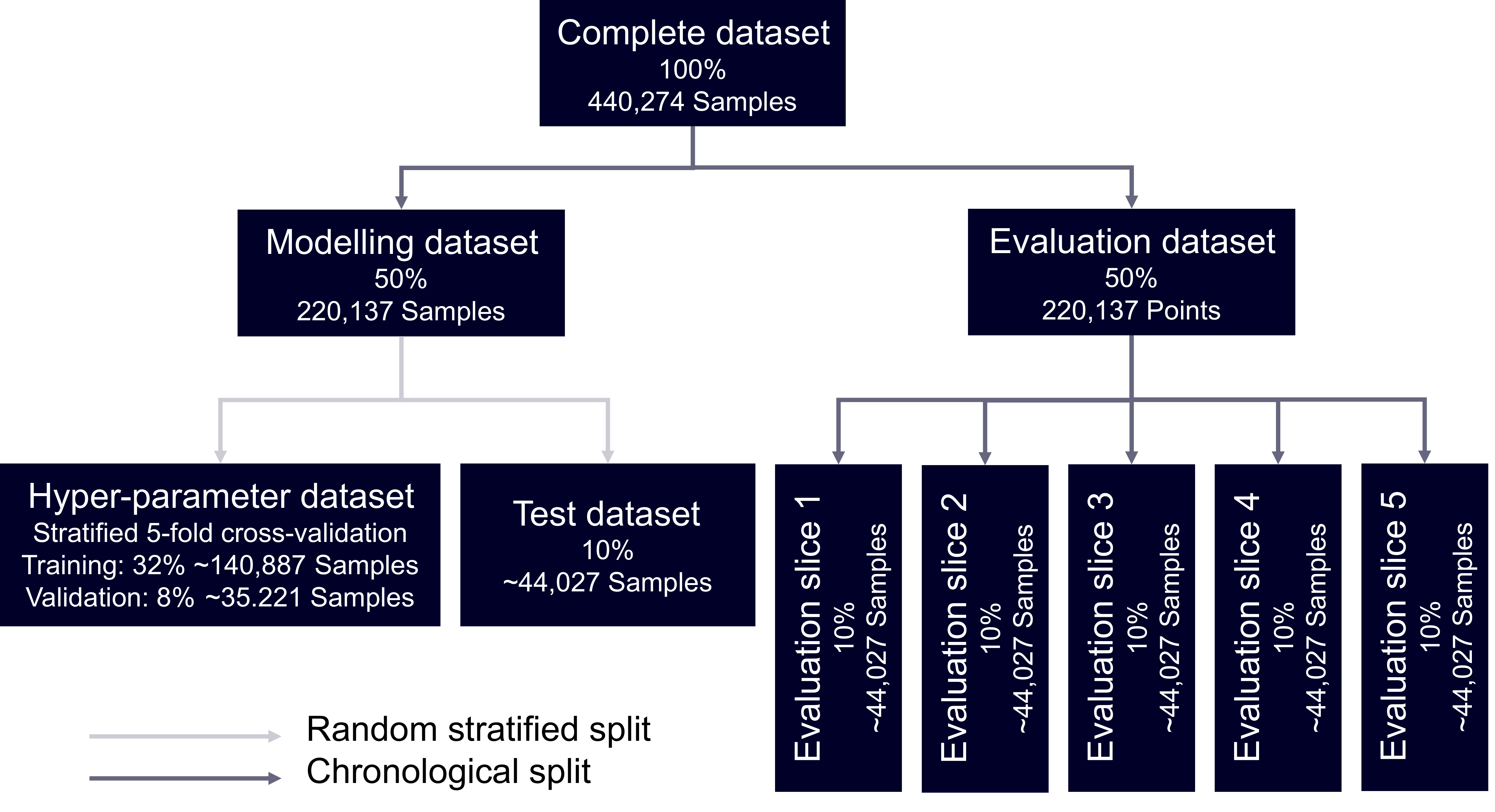}
	  \caption{Data splitting process for our experiment}
   \label{Data_split}
\end{figure}

For our experiment, we initially split our dataset chronologically into two halves. The first half is used for the subsequent modeling. Therefrom, we make a stratified random split in the ratio of 80\% for a hyper-parameter dataset and 20\% for a test dataset. The second half is used after the modeling to evaluate the performance of the model over time by splitting it chronologically into five slices, i.e. 10\% of the total dataset. By this, we can evaluate how the model would perform over five evaluation intervals if it would be deployed to production and the original test dataset has the same size as the evaluation sliced. This logic of splitting our dataset can be seen in Figure \ref{Data_split}, thereby the percent values are related to the total dataset.

With this dataset, we then start a modeling phase. Thereby, we train common models with optimized hyper-parameters determined by Bayesian optimization, whereby in each run of the Bayesian optimization a stratified 5-fold cross-validation on the hyper-parameter dataset is executed. During cross-validation, we evaluate the optimal decision threshold $t_{optimal}$ for each fold. Once we have found proper hyper-parameters, we train once more the model on the whole hyper-parameter dataset and set the decision threshold of the model to the mean value of the optimal decision thresholds from the cross-validation. Then we evaluate the performance of this dataset on the test dataset. By this, we follow the common \ac{ML} modeling scenario with the best practice methods of hyper-parameter optimization and train, validation, and test data splits (cf. Figure \ref{Data_split}). In this modeling, we consider the models \ac{kNN} \cite{Cover1952}, \ac{RFC} \cite{Breiman2001}, \ac{XGBoost} \cite{Chen2016}, and \ac{BRFC} \cite{chen04using}. This procedure is also described in Algorithm \ref{alg:procedure} in Appendix \ref{sec:A}. Additionally, we train a \ac{DC} always predicting the most frequent class and an \ac{AutoML} model based on \cite{feurer-neurips15a} hyper-parameter optimization. The decision threshold of the \ac{AutoML} model is adapted as well.

We execute this modeling phase twice. First, we use standard metrics as the target of the optimization and for evaluation. Second, we use requirement-aware metrics. For more details about the metric definitions see Section \ref{metrics}. Finally, we then evaluate the created models on the evaluation slices to gain insights on how the model would have performed if those models would have been deployed productively. 

As we do not have any benchmark values, we define our research target to find a model that enables a target volume reduction $V_{target} \geq 40\% $ while having a target slip rate $S_{target} \leq 1\%$ as one may assume that slips are much more crucial to the productions than volume reduction. Those values represent realistic requirements from the industrial shopfloor for the use case of false call reduction for \ac{AOI}. Eventually, the \ac{AOI} machine would not be configured that conservatively if it would be another. 

\section{Employed metrics}
\label{metrics}
We use a set of standard metrics and custom requirement-aware metrics for our experiments. We will start in this section to define again the standard metrics first and then come to the custom metrics.  We define positives to be defective boards, and negatives as false calls. For the use case itself, the false negatives, i.e. a defective \ac{PCB} wrongly classified as false call slipping the manual inspection, must be considered much more crucial than false positives, i.e. a non-defective \ac{PCB} boards that are manually inspected, since the latter case just reflects the state of the art situation without \ac{ML} application.

For classification problems, metrics can be grouped by their relationship to a potential decision threshold of the models. Metrics may depend on a set threshold, are independent of a threshold, or give the result on the best possible threshold. 

Common standard metrics that are threshold dependent are accuracy, recall, precision, and F1-score. While accuracy is an often used and widespread metric, it has a strong weakness against imbalanced datasets. For those cases, F1-score is considered often since it is more resilient against those cases and weights recall and precision evenly. 

In comparison to that, different curves for evaluating classifiers can be used for taking different decision thresholds into account, for instance, the \ac{ROC} or \ac{PRC}. As discussed in \cite{Saito2015}, for imbalanced datasets the \ac{PRC} is more informative than the \ac{ROC}. Consequently, we use for this application the \ac{AUC} for this application.

A metric that expresses the trade-off between the recall of two groups is the Youden index \cite{Youden1950}. For our evaluation, we use the best Youden index of a classifier that can be achieved for any decision threshold $t$ and call it Youden score.

In our modeling approach based on standard metrics, we use the metrics accuracy, F1-score, \ac{AUC}, and Youden score as evaluation metrics for model selection. Furthermore, the \ac{AUC} is used as an optimization target and the threshold is set based on the threshold found while calculating the Youden score. By this, we have a mix of commonly used metrics, that either assume a set decision threshold, are completely independent of a decision threshold, or that evaluate the case of the best possible decision threshold.

Even though the named metrics have shown their capabilities for different theoretical research applications, for research on actual applications of \ac{ML} they remain unfit for evaluation purposes. Unlike theoretical research, applications of \ac{ML} always have to justify their cost by achieving certain business criteria to be able to reach a return on investment. However, typically no standard metric reflects those business criteria. Therefore, it is necessary to evaluate \ac{ML} models based on requirement-aware metrics that do directly reflect those business metrics.

The main business metrics for the application of false call reduction are the achieved test volume reduction $v$ (cf. Equation \ref{equ:v}) and the slip rate $s$ (cf. Equation \ref{equ:slip}). Note that in Equation \ref{equ:v} and \ref{equ:slip} $recall_0$ and $recall_1$ refer to the recall metrics for class 0 and class 1, respectively. While $v$ is promising savings for the applications,  $s$ has the danger to produce additional cost. Thus, to identify if the application of false call reduction has positive business impact it is necessary to identify if a model is able to stay below a target slip rate $S_{target}$ and over a target volume reduction $V_{target}$.

\begin{equation}
v = \frac{TN}{TN+FP}=recall_{0}
\label{equ:v}
\end{equation}

\begin{equation}
s = \frac{FN}{TP + FN} = 1-recall_{1}
\label{equ:slip}
\end{equation}

Building onto this, we define three additional metrics that directly reflect the possible business impact that our model might have. As the first metric, we define \ac{cV} as in Equation \ref{equ:cv} with a minimal value of $S_{target}-1$ and a maximal value of one. All negative values of \ac{cV} indicate that the maximum slip rate $S_{target}$ is exceeded with the set threshold.
\begin{equation}
cV = 
\begin{cases}
 S_{target} - s & \text{ if } s \geq S_{target} \\
 v & otherwise
\end{cases}
\label{equ:cv}
\end{equation}
\FloatBarrier 

\begin{figure}[ht]
	\centering
		\includegraphics[width=0.48\textwidth]{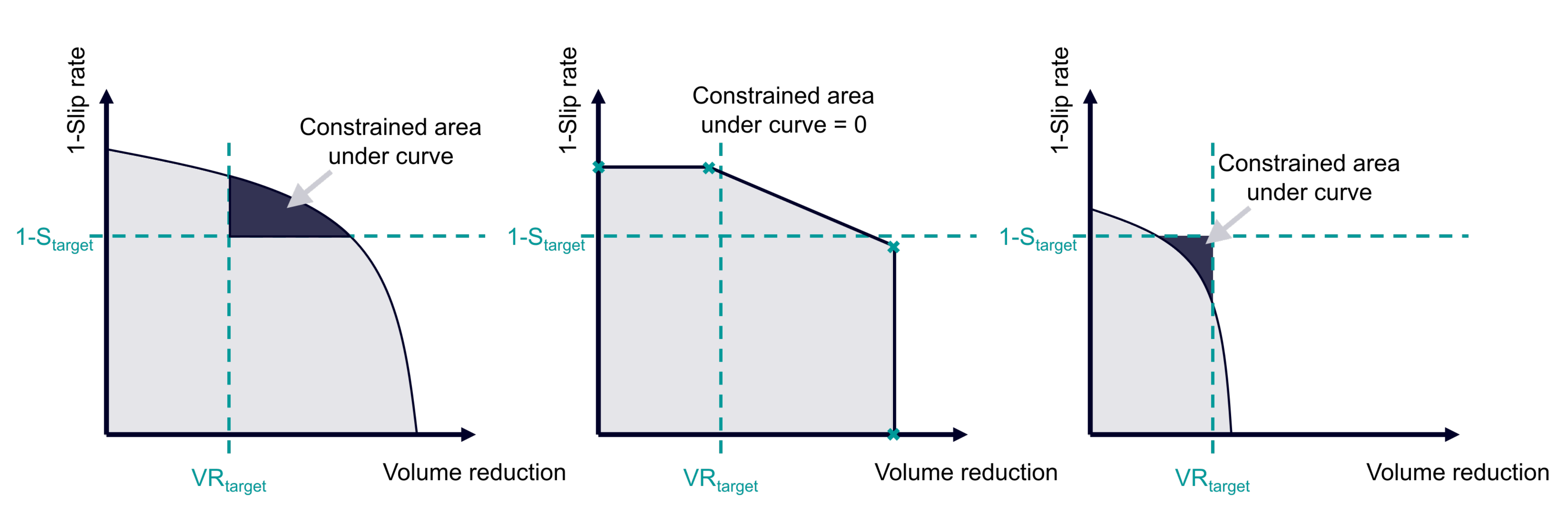}
	  \caption{Visualization of the  \ac{cAUC} metric for case that the classifier has at least one $t$ creating a point in the target zone (left), the case that the classifier has no threshold in the target area but the area under curve intersects with the target area (middle), and the case that the classifier does not intersect with the target area at all (right)}
   \label{fig:cAUC}
\end{figure}

An area-under-curve-based metric is our second requirement-aware metric \ac{cAUC}. In this metric, we express the area under the slip volume reduction curve that lies within the target zone defined by $S_{target}$ and $V_{target}$. The definition of this metric foresees three different cases. For the case where for any $t$ the classifier can fulfill our targets, it has the value of the ratio of the area in the target below the classifier's curve and the target area. In the case where there exists an intersection between the area under the curve and the target zone but in the case where no $t$ the targets are met, the value is zero. The classifier's performance curve is defined by the performance for different decision thresholds and is a discrete function. Thus, the steps of this discrete function can be in such a way that an overlap with the target area is created, even though for all potential thresholds, there is no threshold resulting in a performance within the target area. For the case that there is no intersection of both areas, the metric has the negative area of the gap between the curve and the target zone. Those different cases can also be seen in Figure \ref{fig:cAUC} and are formalized in Equation \ref{equ:cAUC}. Thus, \ac{cAUC} can give values between minus one and one. By this metric, it can be evaluated how well a classifier fulfills for any arbitrary $t$ the business criteria.

\tiny
\begin{equation}
\begin{gathered}
\text{ if } \exists \, t \in [0, 1]: s(t) \leq S_{target}  \land \, v(t) \geq V_{target}: \\
cAUC = \frac{\int_{0}^{1} (1-s)dv - \int_{0}^{V_{target}} (1-s)dv - \int_{0}^{1-S_{target}} (v-V_{target})d(1-s)}{(V_{target}*S_{target})} \\
\text{else if: }\\ \int_{V_{target}}^1 (1-s)dv > \int_{V_{target}}^1 min(1-s,1-S_{target})dv \\
cAUC = 0
\\
\text{otherwise: } \\ cAUC = \frac{\int_{0}^{V_{target}} min(1-s,1-S_{target})dv - (V_{target}*(1-S_{target}))}{(V_{target}*(1-S_{target}))}
\end{gathered}
\label{equ:cAUC}
\end{equation}
\normalsize
As a third requirement-aware metric, we define \ac{V@S} as the maximal volume reduction that fulfills the criteria of falling below the target slip rate for any $t$ (cf. Equation \ref{equ:vs}). By this metric, it is possible to evaluate what volume reduction could be achieved by fulfilling the slip criteria when the perfect decision threshold is set. The metric \ac{V@S} has a minimal value of zero and a maximal value of one. 

\begin{equation}
\begin{gathered}
V@S = \max_{t}(v(t))\quad\forall t \in [0,1] \land s(t) \leq S_{target}
\end{gathered}
\label{equ:vs}
\end{equation}

By those definitions, we again have a metric that is decision threshold specific, a metric that is decision threshold independent and a metric that evaluates the case of the best possible decision threshold. We use all three custom metrics for evaluation metrics for model selection. Furthermore, we use \ac{cAUC} as optimization target and determine the threshold based on the threshold found while calculating \ac{V@S}.

\begin{figure}[ht]
	\centering
		\includegraphics[width=0.48\textwidth]{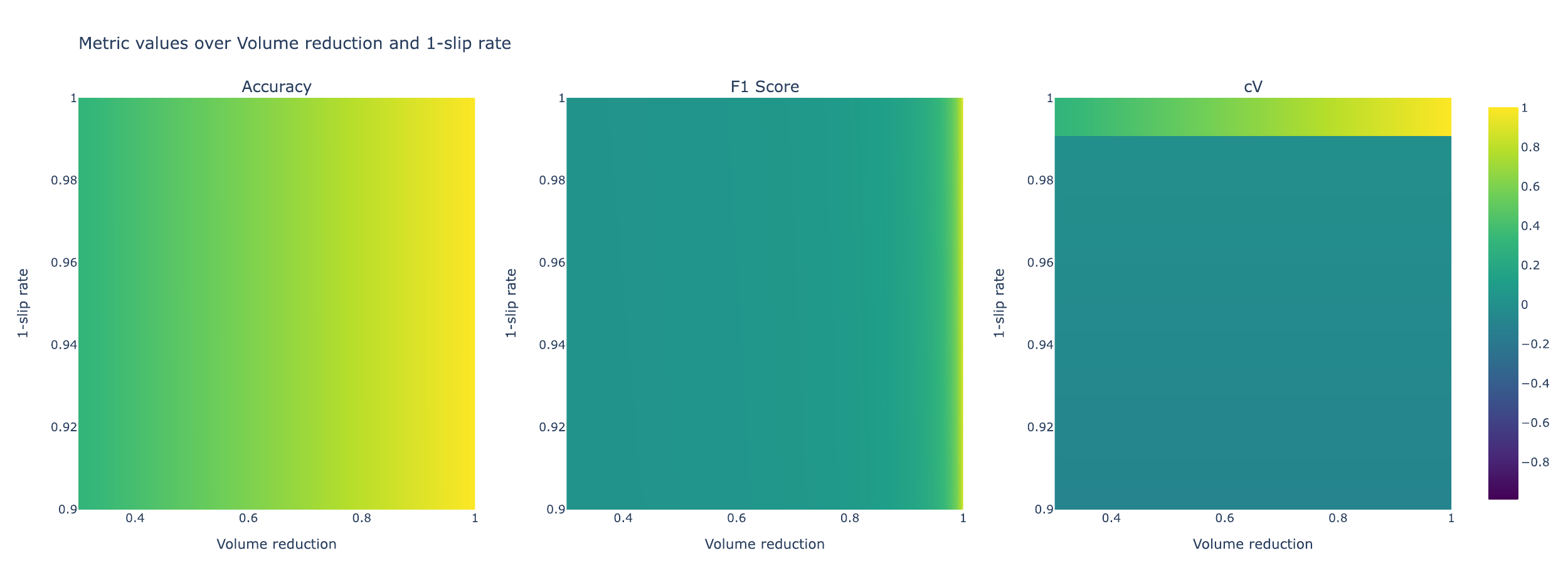}
	  \caption{Visualization of the  metrics Accuracy, F1-Score, and cV for the given dataset's class imbalance.}
   \label{fig:metrics_comp}
\end{figure}

After discussion the standard and custom metrics, we want to give a theoretical comparison of their values for the edge case that one of the targets is exactly not fulfilled and compare the scenario in which the standard metrics are maximal and therefore most misleading. Figure \ref{fig:metrics_comp} shows a comparison for the metrics accuracy, F1-score, and cV for different slip rates and rates of volume reduction. While accuracy overly depends on the volume reduction and therefore has a complete vertical region with values around 1, F1-score is more defines and only gives a smaller vertical area with values around 1. Nevertheless, cV shows only high values in a vertical line on the top right corner, which is indeed aligns with our metric targets. That accuracy is prone to imbalanced data is well known - yet, even F1-Score can reach a value of 0.995 that still does not satisfy the set targets while all of our metric clearly indicates this edge case with values close to 0. Similarly the Youden Score can reach a value up to 0.99 as well as \ac{PRC}. 

Indeed, the defined metrics are customized towards the researched use case. However, the approach of using a set of threshold-dependent, optimal-threshold, and threshold-independent metrics as well as the approach of having requirement-aware metrics can be seen as universal. For our needs, \ac{cAUC} is defined for the slip rate and volume reduction, yet it could also be used for example as a constrained version of the \ac{PRC} or \ac{ROC} curve.

\section{Experimental results}
\label{results}
In this section, we will show and elaborate on the two modeling procedures, whereby one is using standard metrics and the other requirement-aware metrics. For each modeling procedure, we evaluate the created models to rank their performance and select the best ones. After this, we give a comparison with all metrics of the created models to compare the true performance. As the final step, we show the performance of the models on the evaluation slices to investigate their performance if they would have been deployed to the production environment. All results are given as $average \pm standard\:deviation$ of ten runs with different random seeds.

In this first modeling approach, we use standard metrics as discussed in Section \ref{metrics} and adapt the threshold based on the Youden index. The results can be seen in Table \ref{tab:standard_metrics}.

The results show that the model \ac{XGBoost}1 seems to be the best model independently of the metrics. Furthermore, based on the accuracy metric the model \ac{DC}1 seems to outperform all other models except \ac{XGBoost}1. Nevertheless, the other metrics show its poor performance which indicates the in general poor informative value of this metric in regards to imbalanced datasets. In general, accuracy and F1-score indicate poor performance of the  model \ac{BRFC}1 while the Youden score indicates a performance close to the other models. Additionally, the fact that the F1-score of \ac{BRFC}1 is higher than that of the model \ac{kNN}1 while their Youden score has the opposite ranking catches one's eye. This can be explained by the fact that the Youden score evaluates the recalls regarding both classes while the F1-score is considering the precision and the recall in regards to class one combined with the extreme imbalance of the dataset. The model \ac{AutoML}1 is in general performing better than most of the models. Lastly, even though we can analyze and compare the performance of the trained models, it is not possible to conclude whether we have reached our business goals or not as they are not reflected in the available metrics.

\begin{table}[ht]
\centering
\renewcommand{\arraystretch}{1.5}
\caption{Average model performances on the test dataset using standard metrics for ten different random seeds and their standard deviations}
\begin{tabular}{cccccc}
    \toprule
       \makecell{Model\\reference} & Algorithm & Accuracy & F1-score & \ac{AUC} & \makecell{Youden\\score}  \\
    \midrule
        \ac{AutoML}1 & \ac{AutoML} & \makecell{$0.987$\\$\pm0.009$} & \makecell{$0.581$\\$\pm0.176$} & \makecell{$0.866$\\$\pm0.016$} & \makecell{$0.944$\\$\pm0.006$} \\
    \hline
        \ac{DC}1 & \ac{DC} & \makecell{$0.992$\\$\pm0.0$} & \makecell{$0.0$\\$\pm0.0$} & \makecell{$0.504$\\$\pm0.0$} & \makecell{$0.0$\\$\pm0.0$} \\
    \hline
        \ac{BRFC}1 & \ac{BRFC} & \makecell{$0.964$\\$\pm0.005$} & \makecell{$0.299$\\$\pm0.029$} & \makecell{$0.713$\\$\pm0.02$} & \makecell{$0.919$\\$\pm0.013$} \\
    \hline
        \ac{kNN}1 & \ac{kNN} & \makecell{$0.981$\\$\pm0.011$} & \makecell{$0.473$\\$\pm0.146$} & \makecell{$0.739$\\$\pm0.023$} & \makecell{$0.879$\\$\pm0.036$} \\
    \hline
        \ac{RFC}1 & \ac{RFC} & \makecell{$0.983$\\$\pm0.005$} & \makecell{$0.485$\\$\pm0.094$} & \makecell{$0.842$\\$\pm0.016$} & \makecell{$0.948$\\$\pm0.01$} \\
    \hline
        \ac{XGBoost}1 & \ac{XGBoost} & \makecell{$0.993$\\$\pm0.005$} & \makecell{$0.699$\\$\pm0.124$} & \makecell{$0.909$\\$\pm0.015$} & \makecell{$0.959$\\$\pm0.013$} \\
    \bottomrule
\end{tabular}

\label{tab:standard_metrics}
\end{table}

In this first modeling approach, we use  our requirement-aware metrics as discussed in Section \ref{metrics} and adapt the threshold based on the target slip rate. The results can be seen in Table \ref{tab:domain_metrics}.

Based on our requirement-aware metrics  \ac{V@S} and \ac{cAUC} the model \ac{XGBoost}2 seems to perform best. However, the metric \ac{cV} reveals that this model performs much worse than the other models. This pattern indicates, that in general, \ac{XGBoost}2 seems to be a superior model however the method for adapting the threshold works poorly. A similar pattern can be seen for the  model \ac{AutoML}2. Note, that the difference between \ac{AutoML}1 and \ac{AutoML}2 is the target metric used for threshold adaption. Thus, one either must improve the method for adapting the decision threshold first or should instead take the model \ac{BRFC}2 or the model \ac{RFC}2 as those two models are the only ones that on average reach a constrained volume larger than 40\%, i.e. fulfill our set targets for slip rate and volume reduction. However, for both models, their \ac{cV} has a high standard deviation. This might indicate that for the different randoms seeds, not all runs lead to a sufficient model but instead, some models do not fulfill the business requirements and some exceed them significantly. 

\begin{table}[ht]
\centering
\renewcommand{\arraystretch}{1.5}
\caption{Average model performances on the test dataset using requirement-aware metrics for ten different random seeds and their standard deviations}
\begin{tabular}{cccccc}
    \toprule
        \makecell{Model \\reference} & Algorithm & \ac{V@S} & \ac{cV}  & \ac{cAUC}  \\
    \midrule
        \ac{AutoML}2 & \ac{AutoML} & \makecell{$0.906$\\$\pm0.021$} & \makecell{$0.103$\\$\pm0.402$} & \makecell{$0.497$\\$\pm0.136$} \\
    \hline
        \ac{DC}1 & \ac{DC} & \makecell{$0.0$\\$\pm0.0$} & \makecell{$-0.99$\\$\pm0.0$} & \makecell{$-1$\\$\pm0.0$} \\
    \hline
        \ac{BRFC}2 & \ac{BRFC} & \makecell{$0.796$\\$\pm0.064$} & \makecell{$0.423$\\$\pm0.396$} & \makecell{$0.305$\\$\pm0.19$} \\
    \hline
        \ac{kNN}2 & \ac{kNN}& \makecell{$0.478$\\$\pm0.344$} & \makecell{$0.276$\\$\pm0.339$} & \makecell{$0.079$\\$\pm0.091$} \\
    \hline
        \ac{RFC}2 & \ac{RFC} & \makecell{$0.901$\\$\pm0.042$} & \makecell{$0.611$\\$\pm0.416$} & \makecell{$0.5$\\$\pm0.169$} \\
    \hline
        \ac{XGBoost}2 & \ac{XGBoost} & \makecell{$0.925$\\$\pm0.038$} & \makecell{$0.16$\\$\pm0.324$} & \makecell{$0.626$\\$\pm0.12$} \\
    \bottomrule
\end{tabular}

\label{tab:domain_metrics}
\end{table}

\begin{figure}[ht]
	\centering
		\includegraphics[width=0.48\textwidth]{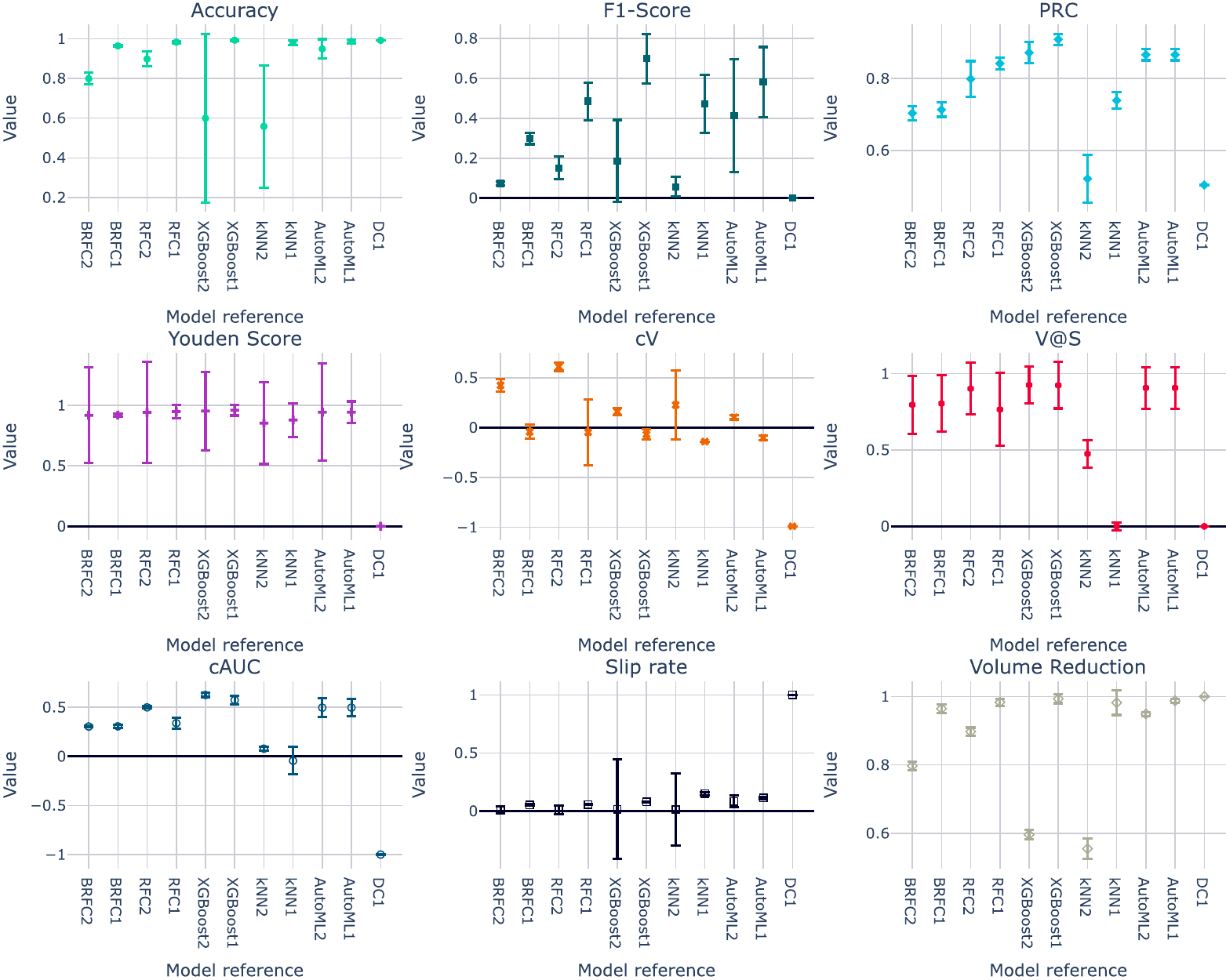}
	  \caption{Performances of all models on the test dataset comparing all metrics}
   \label{model_comparison}
\end{figure}

If we now compare the results of both modeling approaches including all the metrics, we can see in Table \ref{tab:comparison_metrics} and Figure \ref{model_comparison} how the used metrics have impacted our model selection.

\begin{table}[ht]
\centering
\renewcommand{\arraystretch}{1.5}
\caption{Performances of deployable models for all evaluation slices}
\begin{tabular}{cccccccccc}
    \hline
        \multirow{2}{*}{\makecell{Evaluation\\Slice}} & \multicolumn{2}{c}{Model \ac{BRFC}2} & \multicolumn{2}{c}{Model \ac{RFC}2} \\
    \cline{2-5}
         & Slip rate & \makecell{Volume\\Reduction} & Slip rate& \makecell{Volume\\Reduction}\\
    \hline
        1 &  \makecell{$0.073$\\$\pm0.025$} & \makecell{$0.675$\\$\pm0.045$} & \makecell{$0.102$\\$\pm0.074$} & \makecell{$0.667$\\$\pm0.084$}  \\
    \hline
        2 &  \makecell{$0.053$\\$\pm0.016$} & \makecell{$0.487$\\$\pm0.066$} & \makecell{$0.146$\\$\pm0.025$} & \makecell{$0.524$\\$\pm0.122$}  \\
    \hline
        3 &  \makecell{$0.086$\\$\pm0.031$} & \makecell{$0.653$\\$\pm0.066$} & \makecell{$0.15$\\$\pm0.056$} & \makecell{$0.654$\\$\pm0.092$} \\
    \hline
        4 &  \makecell{$0.094$\\$\pm0.01$} & \makecell{$0.78$\\$\pm0.043$} & \makecell{$0.109$\\$\pm0.033$} & \makecell{$0.76$\\$\pm0.083$} \\
    \hline
        5 &  \makecell{$0.093$\\$\pm0.017$} & \makecell{$0.415$\\$\pm0.057$} & \makecell{$0.11$\\$\pm0.046$} & \makecell{$0.397$\\$\pm0.123$} \\
    \hline
\end{tabular}

\label{tab:evaluation_performance}
\end{table}

The results show, that the standard metrics are in many cases contrary to the requirement-aware metrics. For instance, \ac{kNN}2 and DC1 seem to have similar poor behavior based on accuracy, F1-score, and \ac{PRC}, which is connected to the extreme imbalance of the used dataset. Also in general, according to the standard metrics, all models, which hyper-parameters were optimized according to the standard metric \ac{AUC}, seem superior to the models, which hyper-parameters have been optimized according to the requirement-aware metric \ac{cAUC} of the same algorithm. Meanwhile, the requirement-aware metrics show that either the models of the second modeling execution are comparable or better, in specific considering \ac{cV}. Looking at the business metrics, one can see that they are well represented by the requirement-aware metrics but poorly by the standard metrics. 

\begin{figure}[ht]
	\centering
		\includegraphics[width=0.48\textwidth]{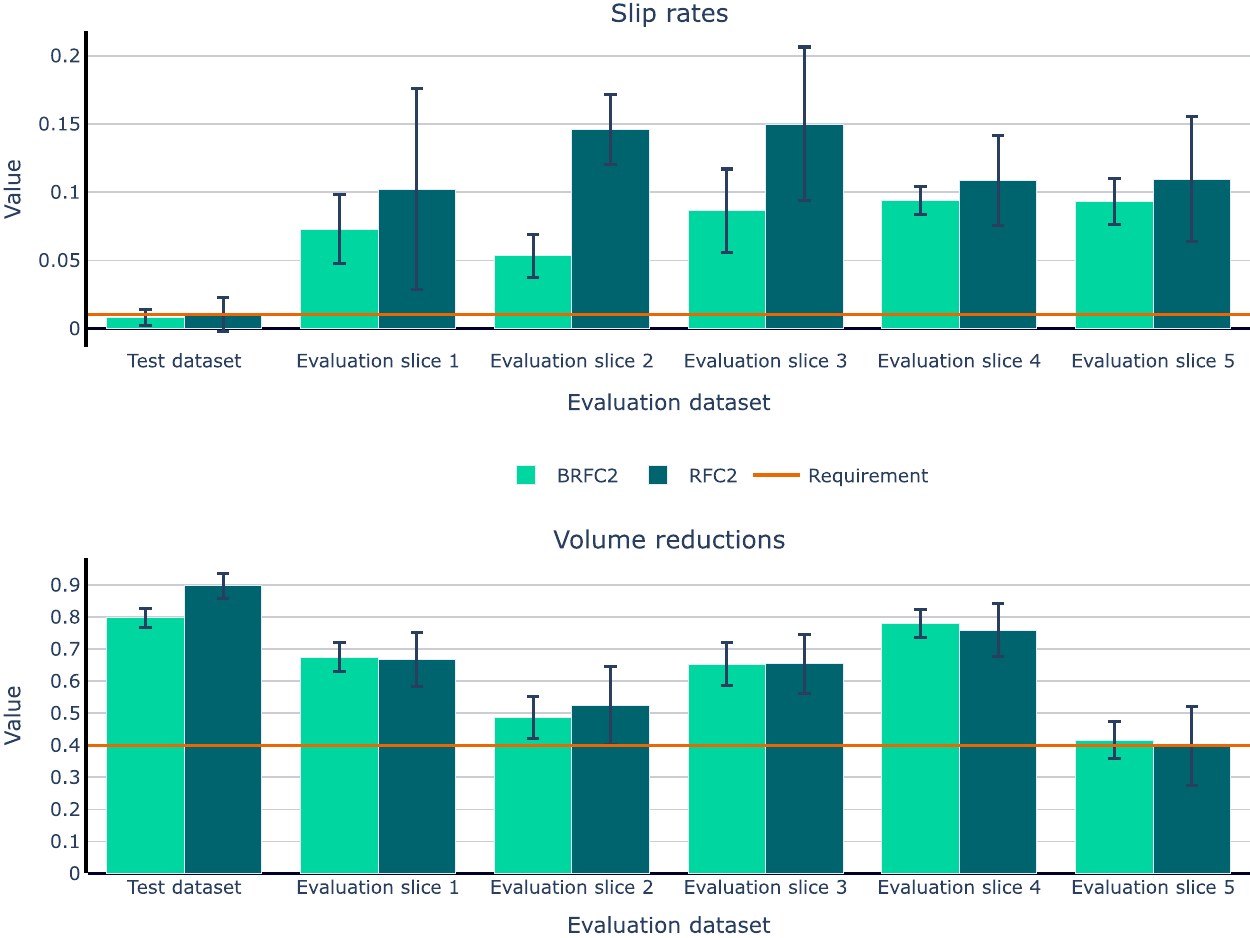}
	  \caption{Performances of deployable models for the test dataset and the evaluation slices}
   \label{timeline}
\end{figure}

The two models \ac{BRFC}2 and \ac{RFC}2 have reached the set targets for slip rate and volume reduction on average. Thus, they now could be deployed to a productive environment. For evaluating the performance over time, we can now use the evaluation slice. Table \ref{tab:evaluation_performance} shows the slip rate and volume reduction for all slices of models \ac{BRFC}2 and \ac{RFC}2. Figure \ref{timeline} offers a visualization of the chronological performance trends.

Even though both models have been promising in the modeling phase, for all evaluation slices the model performances decays immediately. Especially the slip rates exceed already for the first evaluation slice strongly the set target for the slip rate $S_{target}$. The volume reduction decays as well yet the values are still mostly larger than the target $V_{target}$.  Thus, a productive deployment would be a failure and might result in high effort on the shopfloor and cause financial harm. The common evaluation method of a randomly split k-fold cross validation would have failed.

\section{Discussion of results}
\label{discussion}
In this section, we now discuss the effect the different weaknesses defined in Section \ref{methodology} could have had. \ref{W:A} is addressing the reproducibility and transparency of scientific results when not sharing the dataset and the used source code. As our dataset and our source code are openly available \citep{PfabGit, PfabData} it is possible for the scientific community to verify the results but also easily build upon the existing results without the need to have an in detail description of our implementation. The application of anonymization techniques, as for example discussed in \cite{Majeed2021}, can enable the publication of potentially sensitive data. 

Looking at our code base, a large part is the implementation of a Bayesian optimization for different models with different hyper-parameters. Meanwhile, the code for training \ac{AutoML} models is less and simpler. Because of that, \ref{W:B} is addressing the lack of usage of \ac{AutoML} models in related work and instead a strong focus on common models without naming reasons. As by now there exist mature \ac{ML} frameworks making it trivially simple to train models on arbitrary datasets, research should move on and either develop application-specific models, tackle problems that prevent automatizing this aspect of modeling, or focus on problems beyond the modeling. In our results, the AutoML models do not perform best however the performance is absolutely comparable in terms of threshold-independent metrics. For the threshold-dependent metric \ac{cV}, the models performed poorer. This is a consequence of the used method for setting a threshold as it does not work stable enough and has a high standard deviation for the calculated thresholds within the cross-validation. A first benchmark for more selective modeling approaches should be the result of \ac{AutoML} models along with a \ac{DC}.

Regarding \ref{W:C}, our results show clearly the weaknesses of standard metrics, in particular accuracy. As shown in Table \ref{tab:standard_metrics} a \ac{DC} can achieve a better accuracy than properly trained models. Some may argue, that it is common sense that the more imbalanced a dataset is, the more insignificant the metric accuracy is. Yet, it is frequently used in the related works discussed in Section \ref{related_work}. Nevertheless, also other standard metrics are insufficient as they do not reflect the business targets that are met and thus do not allow to classify a modeling procedure to be successful or not. Having proper evaluation metrics is of specific importance when a machine learning operation concept for the application including automatic retraining and model re-deployment is wanted. Also, Table \ref{tab:comparison_metrics} shows clearly that the usage of requirement-aware metrics in the hyper-parameter optimization and the threshold adaption for the model has a crucial impact on successful modeling as standard metrics and application-specific metrics can indicate model performance contrary. Therefore, \ac{ML} research should always first identify the direct business metrics then use evaluation metrics that are directly linked to those business metrics. In the optimal case, those evaluation metrics should be requirement-aware.

\ref{W:D} demands clear targets and requirements in research of \ac{ML} applications that can classify whether an experiment was successful or not. By this, it was possible to select the potential deployable models directly and definitive judge the success of the experiment. Speaking for our results, we have found two models that on average fulfill our set targets on the test data. However, the standard deviation is comparably high and indicates that not all models reach the targets but just the average model. For the evaluation slices, those models fail. Thus, by having defined targets, we can derive now potential next steps to stabilize the modeling by improving the method for setting the threshold and handling the distribution drifts within the dataset. This demonstrates how crucial it is to define clear research targets, as otherwise we could now argue that our research was successful. Especially considering that this is common praxis in other scientific domains, such as statistics, all future applied \ac{ML} research should have clear success criteria defined beforehand. Furthermore, success criteria are the foundations of requirement-aware metrics.

In \ref{W:E} we mention the importance of separating knowledge that is known a priori and knowledge that is only known a posteri. An example can be seen in Table \ref{tab:domain_metrics} and is a matter of setting the threshold. The metric \ac{V@S} uses knowledge that is only available with the ground truth label while \ac{cV} is using only knowledge that is available beforehand by just using the beforehand set threshold. All models have a higher \ac{V@S} than a \ac{cV} even though both metrics actually express both in their positive ranges the volume reduction. In fact, all models of Table \ref{tab:comparison_metrics}, except \ac{kNN}1 and \ac{DC}1, would reach the success criteria according to the metric \ac{V@S}, even though the metric \ac{cV} shows a complete different picture. In a real-world implementation, only a priori information can be used and thus, the expected values are likely to be closer to \ac{cV} than to \ac{V@S}.

The common best practice in \ac{ML} modeling uses a training, validation, and test dataset and assumes that the performance on the test dataset is valid for future data. However, this can be wrong. The used dataset is in its nature tabular and not a time series. So, a random split for training, validating, and testing data is intuitive. However, this should only be done after checking for temporal attributes like distribution drifts within the dataset. There are various distribution drifts happening in this dataset but with the common best practice, researchers risk the potential pitfall of oversee distribution drifts. For instance,  Figure \ref{PCA} would clearly show that there is a time dependency of the dataset and a random split will not lead to realistic performance values. However, this aspect is either not done for academic industrial \ac{AI} research or not discussed (cf. \cite{Jabbar2019, Jamal2022, Lin2006, Thielen2020}). Therefore, \ref{W:F} pleads for an extension of those common best practices also checking for temporal attributes in a dataset before modeling or at least take the chronological last data points as test data set.

For an implementation of an \ac{ML} application, results need to be stable. For this, it is essential to not only evaluate only one run with only one random seed but have multiple runs with different random seeds as pointed out in \ref{W:G}. For instance, for the models \ac{BRFC}2 and \ac{RFC}2 exist runs that have an outstanding low slip rate. However, we can see with multiple runs, that on average the model performances just so reach the targets, but the standard deviation is relatively high. Thus, there is still potential to find more stable methods, for example, by refining the method for setting the threshold.

\section{Conclusion and outlook}
In this work, we discuss the weaknesses of the research methodolgy for the use case of research using false call reduction for \ac{AOI}. We derive seven common weaknesses from peer-reviewed related work and show their consequences experimentally.

In our methodology we explain how we split our dataset allowing a best practice modeling phase but also an analysis of the model performance over time if one would deploy the models. As preparation for our experiment, we define clear success criteria, and define a set of custom requirement-aware metrics that are directly expressing the business impact in contrast to regular metrics such as accuracy, F1-score, or \ac{AUC}. 

Our experiment consists of two different modeling executions using regular metrics and requirement-aware metrics, respectively. For each modeling phase, we select the best model depending on the different metrics to then compare all metrics side by side to demonstrate how misleading inappropriate metrics can be considering the business objectives. Furthermore, we discuss the challenge of setting a proper decision threshold and show the importance of strictly separating between datasets on which the threshold was set and on which the model was evaluated. 

Additionally, we evaluate the performance of the created models over a temporal split evaluation dataset showing that even as sufficient evaluated models fail completely in production. Consequently, we must conclude that in regards of the set success criteria our modelling is successful, but the model would fail from the first moment in production.  

In regards of improving the experimental results, a large contribution to this was done using requirement-aware metrics. While in this work the model type of neural networks has been neglected, the authors \cite{Eban2016} suggest a way to embed performance metrics like precision at recall directly into the loss function used for training the neural network. Further work may dig deeper into extending this approach to a loss function based on the volume reduction on a certain slip rate. 

In the analysis of the model performance over time, one observed a strong performance decay. Thus, it is necessary to create a monitoring system allowing to successfully deploy the model long term. However, for this application, this seems to be in specific a challenge. For monitoring the performance, ground truth labels are required. However, the business case of the application is to no longer generate this ground truth. A naive approach could be using random sampling, but also smarter sampling methods can be found with further research. 

Concluding, the results of our work are useful for two audiences: for researchers focusing on the use case of false call reduction for \ac{AOI} as this is the first work including its source code and its data and for researcher striving for an optimal research methodology for applied \ac{AI} research, as we have shown that the discussed weaknesses are not only present in the related work for false call reduction for \ac{AOI} use, but also in general for use cases in electronic production, and probably beyond.

\bibliographystyle{ACM-Reference-Format}
\bibliography{sample-base}

\FloatBarrier

\appendices

\FloatBarrier
\section{Declaration on the use of AI}
For creating this work, generative AI was used to improve the grammar, conciseness, and clarity of the text. 

\section{Publication acknowledgement}
This paper has been submitted and accept to the IEEE COMPSAC conference 2025. A shorter, 6-pages, version of this paper will be published in the proceedings of the conference.

\section{Procedure for evaluating model performance}
\label{sec:A}
\begin{algorithm}

\SetKwInOut{Input}{Input}
\Input{Random seed, model class, available hyper-parameters}
\KwData{\ac{AOI} dataset}
 \KwResult{Performance metrics for the different stages for a specific model}
 set random seed\;
 split dataset in hyper-parameter dataset, test dataset and evaluation slices\;
 initialize Bayesian optimization\;
 create stratified 5-fold splits from hyper-parameter dataset\;
 \For {20 optimization runs}{
 Retrieve hyper-parameters $hp$ from Bayesian optimization\;
 \For{each split of a stratified 5-fold cross-validation}{
  create model with current $hp$\;
  train model on train folds\;
  calculate model performance $p_{i}$ on validation fold\;
  determine optimal threshold $t_{i}$ on validation fold\;
 }
 calculate average performance $\overline{p}=\frac{\sum_{0}^{5}p_{i}}{5}$ of the target metric\;
 calculate average threshold $\overline{t}=\frac{\sum_{0}^{5}t_{i}}{5}$\;
 update Bayesian optimization\;
 }
 retrieve best hyper-parameter $hp_{opt}$\;
 create model with $hp_{opt}$\;
 train model on hyper-parameter dataset\;
 retrieve and set $\overline{t}$ corresponding to $hp_{opt}$\;
 evaluate model on test dataset and evaluation slices\; 
 \caption{Performance evaluation workflow}
 \label{alg:procedure}
\end{algorithm}

\FloatBarrier
\section{Metric comparison on the test dataset}

\begin{table*}[ht]
\centering
\renewcommand{\arraystretch}{1.1}
\caption{Performances of all models on the test dataset showing all metrics}
\begin{tabular}{cccccccccc}
    \toprule
        \multirow{2}{*}{\makecell{Model \\reference}}&\multicolumn{4}{c}{Standard metrics} & \multicolumn{3}{c}{Requirement-aware metrics} & \multicolumn{2}{c}{Business metrics}\\
    \cline{2-10}
         & Accuracy & F1-score &  \ac{PRC} & \makecell{Youden\\score}  &  \ac{V@S} & \ac{cV} & \ac{cAUC} & Slip rate& \makecell{Volume\\Reduction}\\
    \midrule
        \ac{AutoML}1 &  \makecell{$0.987$\\$\pm0.009$} & \makecell{$0.581$\\$\pm0.176$} & \makecell{$0.866$\\$\pm0.016$} & \makecell{$0.944$\\$\pm0.006$} & \makecell{$0.906$\\$\pm0.021$} & \makecell{$-0.103$\\$\pm0.089$} & \makecell{$0.497$\\$\pm0.136$} & \makecell{$0.113$\\$\pm0.089$} & \makecell{$0.988$\\$\pm0.01$}\\
    \hline
        \ac{AutoML}2 &  \makecell{$0.949$\\$\pm0.049$} & \makecell{$0.413$\\$\pm0.283$} & \makecell{$0.866$\\$\pm0.016$} & \makecell{$0.944$\\$\pm0.006$} & \makecell{$0.906$\\$\pm0.021$} & \makecell{$0.103$\\$\pm0.402$} & \makecell{$0.497$\\$\pm0.136$} & \makecell{$0.083$\\$\pm0.095$} & \makecell{$0.949$\\$\pm0.049$}\\
    \hline
        DC1 & \makecell{$0.992$\\$\pm0.0$} & \makecell{$0.0$\\$\pm0.0$} & \makecell{$0.504$\\$\pm0.0$} & \makecell{$0.0$\\$\pm0.0$} & \makecell{$0.0$\\$\pm0.0$} & \makecell{$-0.99$\\$\pm0.0$} & \makecell{$-1$\\$\pm0.0$} & \makecell{$1.0$\\$\pm0.0$} & \makecell{$1.0$\\$\pm0.0$}\\
    \hline
        \ac{BRFC}1  & \makecell{$0.964$\\$\pm0.005$} & \makecell{$0.299$\\$\pm0.029$} & \makecell{$0.713$\\$\pm0.02$} & \makecell{$0.919$\\$\pm0.013$} & \makecell{$0.804$\\$\pm0.072$} & \makecell{$-0.042$\\$\pm0.015$} & \makecell{$0.307$\\$\pm0.185$} & \makecell{$0.052$\\$\pm0.015$} & \makecell{$0.964$\\$\pm0.005$}\\
    \hline
        \ac{BRFC}2  & \makecell{$0.799$\\$\pm0.03$} & \makecell{$0.074$\\$\pm0.01$} & \makecell{$0.704$\\$\pm0.019$} & \makecell{$0.918$\\$\pm0.012$} & \makecell{$0.796$\\$\pm0.064$} & \makecell{$0.423$\\$\pm0.396$} & \makecell{$0.305$\\$\pm0.19$} & \makecell{$0.008$\\$\pm0.006$} & \makecell{$0.797$\\$\pm0.03$}\\
    \hline
        \ac{kNN}1 & \makecell{$0.981$\\$\pm0.011$} & \makecell{$0.473$\\$\pm0.146$} & \makecell{$0.739$\\$\pm0.023$} & \makecell{$0.879$\\$\pm0.036$} & \makecell{$0.0$\\$\pm0.0$} & \makecell{$-0.14$\\$\pm0.139$} & \makecell{$-0.044$\\$\pm0.026$} & \makecell{$0.15$\\$\pm0.139$} & \makecell{$0.982$\\$\pm0.012$}\\
    \hline
        \ac{kNN}2 & \makecell{$0.558$\\$\pm0.308$} & \makecell{$0.057$\\$\pm0.049$} & \makecell{$0.521$\\$\pm0.067$} & \makecell{$0.853$\\$\pm0.031$} & \makecell{$0.475$\\$\pm0.344$} & \makecell{$0.227$\\$\pm0.339$} & \makecell{$0.079$\\$\pm0.091$} & \makecell{$0.014$\\$\pm0.018$} & \makecell{$0.555$\\$\pm0.311$}\\
    \hline
        \ac{RFC}1 & \makecell{$0.983$\\$\pm0.005$} & \makecell{$0.485$\\$\pm0.094$} & \makecell{$0.842$\\$\pm0.016$} & \makecell{$0.948$\\$\pm0.01$} & \makecell{$0.765$\\$\pm0.328$} & \makecell{$-0.047$\\$\pm0.057$} & \makecell{$0.339$\\$\pm0.239$} & \makecell{$0.057$\\$\pm0.057$} & \makecell{$0.983$\\$\pm0.005$}\\
    \hline
        \ac{RFC}2 & \makecell{$0.898$\\$\pm0.038$} & \makecell{$0.015$\\$\pm0.057$} & \makecell{$0.799$\\$\pm0.049$} & \makecell{$0.942$\\$\pm0.012$} & \makecell{$0.901$\\$\pm0.042$} & \makecell{$0.611$\\$\pm0.416$} & \makecell{$0.5$\\$\pm0.169$} & \makecell{$0.01$\\$\pm0.012$} & \makecell{$0.897$\\$\pm0.039$}\\
    \hline
        \ac{XGBoost}1 & \makecell{$0.993$\\$\pm0.005$} & \makecell{$0.699$\\$\pm0.124$} & \makecell{$0.909$\\$\pm0.015$} & \makecell{$0.959$\\$\pm0.013$} & \makecell{$0.923$\\$\pm0.051$} & \makecell{$-0.068$\\$\pm0.043$} & \makecell{$0.574$\\$\pm0.152$} & \makecell{$0.078$\\$\pm0.043$} & \makecell{$0.993$\\$\pm0.005$}\\
    \hline
        \ac{XGBoost}2 & \makecell{$0.599$\\$\pm0.425$} & \makecell{$0.185$\\$\pm0.206$} & \makecell{$0.872$\\$\pm0.029$} & \makecell{$0.953$\\$\pm0.013$} & \makecell{$0.923$\\$\pm0.051$} & \makecell{$0.16$\\$\pm0.324$} & \makecell{$0.626$\\$\pm0.12$} & \makecell{$0.015$\\$\pm0.021$} & \makecell{$0.596$\\$\pm0.429$}\\
    \bottomrule
\end{tabular}

\label{tab:comparison_metrics}
\end{table*}
\FloatBarrier

\end{document}